\documentclass{article}

\usepackage[sglblindworkshop, final]{neurips_2025}

\usepackage[utf8]{inputenc} 
\usepackage[T1]{fontenc}    
\usepackage{hyperref}       
\usepackage{url}            
\usepackage{booktabs}       
\usepackage{amsfonts}       
\usepackage{nicefrac}       
\usepackage{microtype}      
\usepackage{xcolor}         
\usepackage{graphicx}
\usepackage{subcaption}
\usepackage[font=small,labelfont=bf]{caption} 
\usepackage{wrapfig}
\usepackage{caption}
\usepackage[most]{tcolorbox}
\tcbset{
  colback=blue!5,    
  colframe=blue!40,  
  boxrule=0.5pt,
  arc=2pt,
  left=6pt,
  right=6pt,
  top=4pt,
  bottom=4pt
}
\graphicspath{{Figures/}} 

\usepackage{multirow}

\title{From Snow to Rain: Evaluating Robustness, Calibration, and Complexity of Model-Based Robust Training}

\author{%
  Josué Martínez-Martínez\thanks{Work completed under summer research program at MIT Lincoln Laboratory.}, \ Olivia Brown, Giselle Zeno,\\
  \textbf{Pooya Khorrami, Rajmonda Caceres}
\\
  MIT Lincoln Laboratory \\
  Lexington, MA  \\
  \texttt{\{josue.martinez-martinez, olivia.brown, giselle.zeno,}\\
  \texttt{pooya.khorrami, rajmonda.caceres\}@ll.mit.edu}
}

\begin{document}

\maketitle

\begin{abstract}
Robustness to natural corruptions remains a critical challenge for reliable deep learning, particularly in safety-sensitive domains. 
We study a family of model-based training approaches that leverage a learned nuisance variation model to generate realistic corruptions, as well as new hybrid strategies that combine random coverage with adversarial refinement in nuisance space. 
Using the Challenging Unreal and Real Environments for Traffic Sign Recognition dataset (CURE-TSR), with Snow and Rain corruptions, we evaluate accuracy, calibration, and training complexity across corruption severities. 
Our results show that model-based methods consistently outperform baselines Vanilla, Adversarial Training, and AugMix baselines, with model-based adversarial training providing the strongest robustness under across all corruptions but at the expense of higher computation and model-based data augmentation achieving comparable robustness with $T$ less computational complexity without incurring a statistically significant drop in performance. 
These findings highlight the importance of learned nuisance models for capturing natural variability, and suggest a promising path toward more resilient and calibrated models under challenging conditions.
\end{abstract}

\vspace{-5mm}
\section{Introduction}
\vspace{-3mm}
Snow on a traffic sign, rain on a windshield, or blur from a camera shake---these everyday corruptions are enough to make state-of-the-art deep networks fail. 
Such failures are not only common but also dangerous in safety-critical domains such as autonomous driving, where robust perception is essential. 
Corruption benchmarks~\cite{hendrycks2019benchmarkingneuralnetworkrobustness} demonstrate that models trained only on clean data can lose more than half their accuracy under natural corruptions. 
Even worse, these models often remain highly confident in their incorrect predictions, creating a serious reliability gap.

A widely used strategy to address robustness is data augmentation with analytical transformations. 
Techniques such as flips, crops, brightness jitter, or composition methods like AugMix~\cite{hendrycks2019augmix} expose the model to synthetic diversity during training. 
These augmentations improve robustness in the average case and are computationally cheap. 
However, because they rely on handcrafted transformations, they cannot capture the full richness of natural variability and often fail under severe corruptions.

Model-based augmentation offers a more powerful alternative by learning nuisance variation directly from data~\cite{robey2020model}. 
Instead of relying on predefined rules, these methods use generative models to capture realistic transformations observed in real corrupted images. 
This makes them particularly well-suited for natural corruptions, since the generator can reproduce features like snow texture or rain streaks that analytical methods cannot simulate convincingly. 
As a result, model-based approaches can produce training data that better reflects real-world distributions, improving both robustness and calibration under challenging conditions.

In this work, we conduct a systematic analysis of model-based augmentation methods together with hybrid variants that combine random sampling or worst-of-$k$ selection with adversarial refinement in nuisance space.  
We place these methods alongside analytical baselines on the Challenging Unreal and Real Environments for Traffic Sign Recognition (CURE-TSR)~\cite{temel2018curetsrchallengingunrealreal} under Snow and Rain corruptions, evaluating both classification accuracy and calibration across severities.  
In addition, we analyze their training complexity, providing insight into the trade-offs between robustness, calibration, and efficiency.  
Our study extends prior evaluations of hybrids on simple corruptions such as brightness and contrast, and offers the first systematic comparison under realistic natural corruptions.

\paragraph{Contributions.}
Our work makes three contributions:
(1) we provide a systematic comparison of analytical augmentations (Vanilla, AugMix) and model-based methods (model-based data augmentation (MDA), model-based robust training (MRT), model-based adversarial training (MAT)), together with hybrid variants, under Snow and Rain corruptions in CURE-TSR; 
(2) we evaluate both robustness and calibration, providing insight into how different augmentation strategies trade off accuracy and confidence under natural corruptions; 
(3) we perform an analysis on computational complexity, clarifying the efficiency costs of random, adversarial, and hybrid nuisance generation.

\section{Related Work}

\paragraph{Analytical augmentations.}
Data augmentation has long been a standard tool for improving generalization and robustness in deep learning, and remains one of the simplest and most widely adopted practices across domains.  
Classical image transformations such as flips, rotations, crops, and color jitter increase training diversity at almost no computational cost, reducing overfitting and improving baseline robustness~\cite{deepresidual2016}.  
Over time, more advanced analytical methods have been proposed to further diversify synthetic examples.  
MixUp~\cite{zhang2018mixupempiricalriskminimization} interpolates pairs of images and their labels, encouraging models to behave linearly in-between training samples.  
CutMix~\cite{yun2019cutmixregularizationstrategytrain} pastes patches from one image into another, forcing networks to learn from partial and mixed visual evidence.  
PixMix~\cite{hendrycks2022pixmixdreamlikepicturescomprehensively} combines natural images with perturbations drawn from unrelated datasets, substantially increasing sample diversity.  
AugMix~\cite{hendrycks2019augmix} further improves robustness by composing multiple augmentations and enforcing prediction consistency across augmented views, yielding strong gains on corruption benchmarks.  
These methods are computationally efficient, easy to implement, and highly effective in average-case robustness, which has made them a default component in modern training pipelines.  
However, they all rely on handcrafted transformations and pixel-level heuristics that are not designed to mimic the structured, high-level variability of natural corruptions such as snow, rain, or fog.  
As a result, analytical augmentations tend to generalize poorly to unseen corruptions and often fail under severe distribution shifts, motivating the exploration of model-based alternatives.

\paragraph{Model-based augmentations.}
A growing line of work leverages generative models to simulate nuisance variability directly from data, enabling augmentations that more faithfully reflect real-world corruption structure.  
Foundational generative families include VAEs~\cite{kingma2022autoencodingvariationalbayes}, GANs~\cite{goodfellow2014generativeadversarialnetworks}, and disentangling frameworks such as MUNIT~\cite{huang2018multimodalunsupervisedimagetoimagetranslation} and DRIT~\cite{lee2019dritdiverseimagetoimagetranslation}.  
These models have been applied in domain adaptation and distribution shift mitigation (e.g., pixel-level translation with cycle consistency)~\cite{hoffman2017cycadacycleconsistentadversarialdomain}, and in medical imaging, where learned augmentations improve cross-device and cross-protocol generalization~\cite{bowles2018ganaugmentationaugmentingtraining}.  
More recently, Robey et al.~\cite{robey2020model} formalized model-based robust training techniques such as, MDA, MRT, MAT, showing that augmentations driven by a learned variation model can improve robustness and calibration under corruption benchmarks.  
Compared to analytical methods, model-based approaches have the advantage of capturing realistic patterns (e.g., snow texture, rain streaks, lighting shifts) directly from data, enabling them to produce training examples that better reflect real-world variability.  
However, they also introduce higher computational cost and create new design choices around how to balance random coverage, worst-case sampling, and adversarial optimization in nuisance space.

\paragraph{Hybrid variants.}
Beyond purely random or purely adversarial nuisance generation, hybrid approaches seek to combine broad coverage with targeted refinement.  
At the policy level, automated augmentation techniques like AutoAugment~\cite{cubuk2019autoaugmentlearningaugmentationpolicies}, RandAugment~\cite{cubuk2019randaugmentpracticalautomateddata}, and Adversarial AutoAugment~\cite{zhang2019adversarialautoaugment} discover hard augmentation policies via search or adversarial bilevel optimization.  
From a robustness standpoint, methods in adversarial data augmentation and distributionally robust optimization (DRO) formalize worst-case sampling over perturbation sets~\cite{volpi2018generalizingunseendomainsadversarial}, with practical strategies using multi-sample restarts or worst-of-$k$ selection~\cite{uesato2018adversarialriskdangersevaluating}.  
Hybrid variants have also been explored in prior work that combined model-based random sampling with adversarial refinement, or worst-of-$k$ selection with adversarial refinement, and compared them to analytical approaches under simple corruptions such as brightness and contrast~\cite{martinez2024towards}.  
These studies found that analytical techniques like AugMix remained competitive or superior on accuracy in such cases, while MAT often achieved the lowest calibration error.  
However, the performance of these hybrids under more realistic natural corruptions (e.g., snow, rain) has not been systematically evaluated.  
In this work, we place hybrid strategies alongside analytical and model-based baselines, providing a unified comparison of their robustness, calibration, and complexity trade-offs.

\vspace{-2mm}
\section{Methodology}
We evaluate training strategies that target robustness to natural corruptions via either input-level augmentations or model-based variations.
Our aim is to compare their robustness and calibration, and to expose the computational trade-offs that arise during training.

\paragraph{Standard baselines.}
We employ three methods that do not leverage a learned model of natural variation to serve as our baselines.
We first consider \textbf{Vanilla} empirical risk minimization (ERM), trained only on clean data.
We also evaluate \textbf{Adversarial Training (AT)}, which defends against additive perturbations using projected gradient descent (PGD) in input space~\cite{madry2017towards}.
Finally, we include \textbf{AugMix}~\cite{hendrycks2019augmix}, which induces synthetic diversity by mixing multiple analytical augmentations.

\paragraph{Model-based approaches.}


We used a learned variation model $G(x,z)$ to capture nuisance transformations $z$ applied to input $x$~\cite{robey2020model}. 
Formally, the model-based objective is given in Equation~\eqref{eq:modelbased}:
\vspace{-6mm}

\begin{equation}
\min_{\theta}\;
\mathbb{E}_{(x,y)\sim\mathcal{D}}
\Big[
\max_{z\in\mathcal{Z}}
L\big(f_{\theta}(G(x,z)),y\big)
\Big],
\label{eq:modelbased}
\end{equation}
where $\mathcal{D}$ is the data distribution and $\mathcal{Z}$ denotes the nuisance space.
In practice we bound the nuisance space by a \emph{severity} hyperparameter $\rho$ and write $\mathcal{Z}_\rho$, which controls the magnitude of allowable variation during training.
We study three approximations to the inner maximization:
\begin{itemize}
\item \textbf{Model-based Data Augmentation (MDA).} Replace $\max_{z\in\mathcal{Z}_\rho}$ with a single random draw $z\sim\mathcal{Z}_\rho$ per sample.
\item \textbf{Model-based Robust Training (MRT).} Sample $k$ candidates $\{z_i\}$ from $\mathcal{Z}_\rho$ and select the one with the highest loss.
\item \textbf{Model-based Adversarial Training (MAT).} Perform PGD in $z$ with $T$ steps and projection onto $\mathcal{Z}_\rho$ to approximate the inner maximum.
\end{itemize}

\paragraph{Hybrid methods.}
They combine coverage from random or sampled nuisances with the sharpness of adversarial refinement in nuisance space.
\begin{itemize}
\item \textbf{MDAT (MDA $\rightarrow$ MAT).} Initialize MAT’s PGD with a single random $z\!\sim\!\mathcal{Z}_\rho$ drawn by MDA, then run $T$ adversarial steps in $z$.
\item \textbf{MRAT (MRT $\rightarrow$ MAT).} Run MRT for $k$ trials to identify the highest-loss $z$ and use it to initialize MAT’s $T$-step PGD in $z$.
\end{itemize}
These hybrids are designed to balance broad nuisance coverage and targeted stress testing, yielding diverse yet challenging training examples.
\vspace{-3mm}
\paragraph{Complexity summary.}
Table~\ref{tab:complexity} summarizes per-batch training cost.
MDA and AugMix are $\Theta(1)$ but incur larger constants than Vanilla due to additional generator or augmentation calls.
MRT scales as $\Theta(k)$ with the number of nuisance trials, and MAT scales as $\Theta(T)$ with the number of PGD steps in $z$.
MDAT inherits MAT’s cost, $\Theta(T)$, while MRAT composes both to $\Theta(k{+}T)$.
At inference time all methods run the same classifier $f_\theta$ on the input, so the evaluation cost is unchanged.


\begin{table}[t]
\centering
\small
\setlength{\tabcolsep}{6pt}
\caption{\textbf{Training complexity per mini-batch.} 
Notation: $k$ = number of nuisance samples in MRT, $T$ = number of PGD iterations in nuisance space (MAT), $R$ = number of PGD iterations in input space (AT). 
The key distinction is that AT scales with $\Theta(R)$ due to input-space PGD, while MAT scales with $\Theta(T)$ due to nuisance-space PGD. 
Hybrids inherit the cost of their components (MDAT $\approx \Theta(T)$, MRAT $\approx \Theta(k+T)$).}
\begin{tabular}{lll}
\toprule
\textbf{Method} & \textbf{Inner-loop work per batch} & \textbf{Asymptotic cost} \\
\midrule
Vanilla & One forward/backward pass & $\Theta(1)$ \\
AT (PGD) & $R$ gradient steps in input space & $\Theta(R)$ \\
AugMix & Multiple random augmentations & $\Theta(1)$ (larger const.) \\
MDA & One random nuisance $z$ sample via generator & $\Theta(1)$ (larger const.) \\
MRT & Nuisance search for $k$ trials & $\Theta(k)$ \\
MAT & PGD steps in $z$ for $T$ steps & $\Theta(T)$ \\
MDAT & 1 random $z$ seed + MAT for $T$ steps & $\Theta(T)$ \\
MRAT & MRT for $k$ steps + MAT for $T$ steps & $\Theta(k+T)$ \\
\bottomrule
\end{tabular}

\vspace{-5mm}
\label{tab:complexity}
\end{table}

\section{Experimental Setup}


\paragraph{Dataset.}
We evaluate on CURE-TSR~\cite{temel2018curetsrchallengingunrealreal}, a standard dataset for multi-class image classification containing 14 classes of traffic signs and approximately 2.2 million images in total. 
The dataset comprises both real-world and synthetically generated samples and includes 12 types of challenging conditions (e.g., blur, contrast changes, weather, noise, and occlusion), each applied at five severity levels. 
All images are resized to 32 × 32. 
For our experiments, we used the 19,610 challenge-free images for training, and evaluated on two natural corruptions from this benchmark, Snow and Rain, each comprising 3,334 test images per severity level (1–5), in addition to the clean test data (severity 0). 
Severity 1 corresponds to mild degradation, while severity 5 represents extreme corruption.

\paragraph{Metrics.}
We report two complementary metrics.
First, classification \textbf{accuracy} measures robustness under corruption.
Second, \textbf{Expected Calibration Error (ECE)}~\cite{kumar2020verifieduncertaintycalibration} evaluates whether model confidence is aligned with predictive accuracy.
ECE is computed using equally sized bins.
All reported results are averaged over five independent training runs with different random seeds, and we report mean~$\pm$~standard deviation (std).


\paragraph{Model architecture.}
Following Robey et al.~\cite{robey2020model}, we adopt their convolutional neural network classifier, which consists of two convolutional layers followed by two fully connected layers.
In addition, the variation model $G(x,z)$ used in MDA, MRT, MAT, and the hybrid approaches is the MUNIT-style autoencoder~\cite{huang2018multimodalunsupervisedimagetoimagetranslation} introduced in their work, trained to reconstruct and vary traffic sign images under realistic nuisance factors.

\begin{figure*}[t]
  \centering
  \begin{subfigure}{\textwidth}
    \centering
    \includegraphics[width=\linewidth]{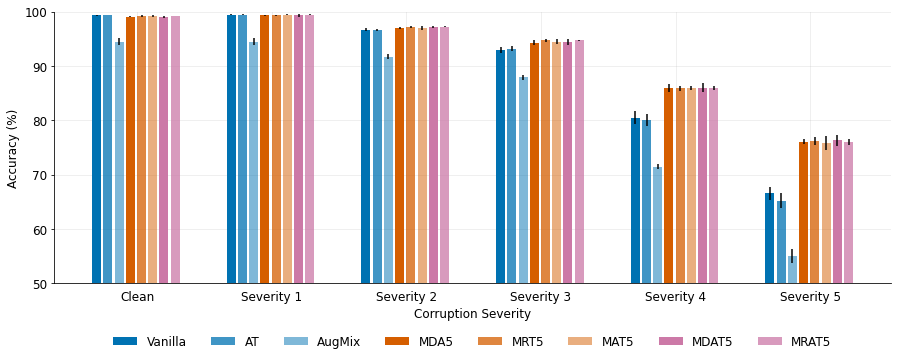}
    \caption{Accuracy (mean $\pm$ std) vs. severity.}
    \label{fig:gtsrb-snow-acc}
  \end{subfigure}

  \vspace{1em} 

  \begin{subfigure}{\textwidth}
    \centering
    \includegraphics[width=\linewidth]{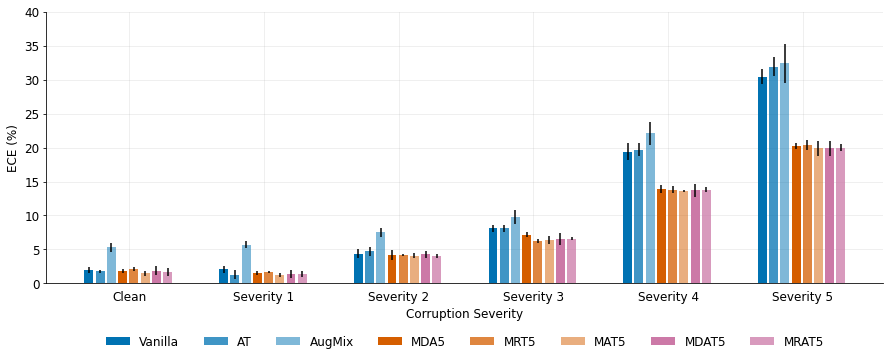}
    \caption{ECE (mean $\pm$ std) vs. severity.}
    \label{fig:gtsrb-snow-ece}
  \end{subfigure}

  \caption{\textbf{CURE-TSR under Snow corruption.} 
  Vanilla and AT degrade quickly with severity, while AugMix collapses even more sharply. 
  Model-based methods (MDA, MRT, MAT) sustain substantially higher robustness and lower ECE, retaining $\sim 75\%$ accuracy and $\sim 20\%$ ECE at severity~5 compared to $\sim 55$--$65\%$ and $30$--$32\%$ for baselines.}
  \label{fig:gtsrb-snow-side}
  \vspace{-4mm}
\end{figure*}

\paragraph{Training setup.}
All models are trained for 100 epochs using Adadelta with a learning rate of $1.0$.
We use a batch size of $64$ across all methods to ensure comparability.
For AT, we use additive perturbations bounded by $\ell_\infty$ norm of size $8/255$, with a step size of $0.01$, and $10$ iterations of projected gradient descent.
For MRT we use $k$ nuisance samples per example, and for MAT we use $T$ projected gradient ascent steps in nuisance space.
Hybrids MDAT and MRAT follow their respective initialization strategies before running MAT.
Unless otherwise specified, we report results with $k = 10$ and $T = 10$.
Models are trained at two different nuisance severities (levels 2 and 5), while evaluation is performed across the full range of severities.
In our training pipeline, we augment each challenge-free image with a corrupted counterpart, generated either by the variation model (for nuisance corruptions), by the PGD algorithm (for adversarial perturbations), or by AugMix (for randomized analytical augmentations); the only exception is the Vanilla method, which uses challenge-free images alone. 
For all methods, we optimize a cross-entropy loss on both the clean and corrupted images and sum the two losses. 
In the case of AugMix, we also follow this formulation rather than using the Jensen–Shannon loss proposed in the original study~\cite{hendrycks2019augmix}.


\section{Results}
In the following, we denote each method by appending the severity level used during training to its name (e.g., MDA2 for training at severity 2, and MDA5 for training at severity 5).
\begin{figure*}[t]
  \centering
  \begin{subfigure}{\textwidth}
    \centering
    \includegraphics[width=\linewidth]{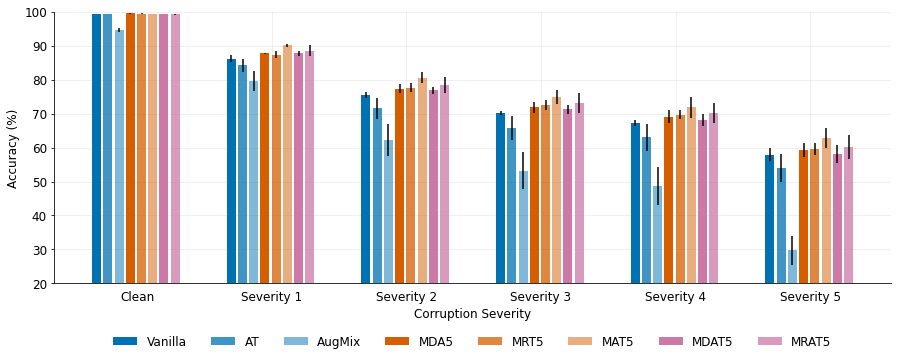}
    \caption{Accuracy (mean $\pm$ std) vs. severity.}
    \label{fig:gtsrb-rain-acc}
  \end{subfigure}
  \hfill
  \begin{subfigure}{\textwidth}
    \centering
    \includegraphics[width=\linewidth]{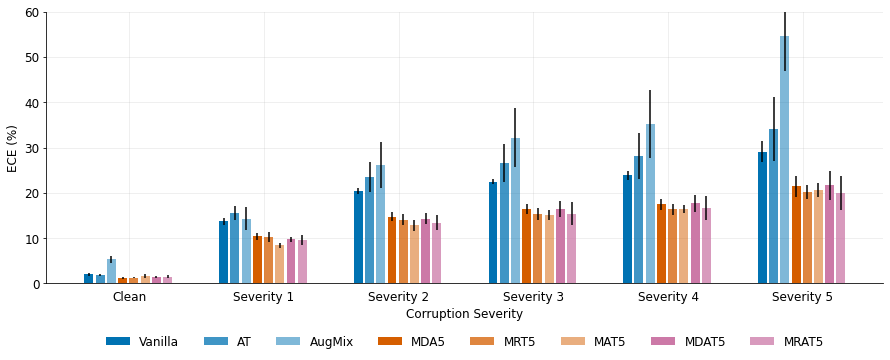}
    \caption{ECE (mean $\pm$ std) vs. severity.}
    \label{fig:gtsrb-rain-ece}
  \end{subfigure}
  \vspace{-7mm}
\caption{\textbf{CURE-TSR under Rain corruption.} 
Rain is more challenging than Snow, causing Vanilla, AT, and AugMix to collapse below $60\%$ accuracy and above $30\%$ ECE at severity~5. 
Model-based methods (MDA, MRT, MAT) markedly improve robustness: MRT5 and MDA5 reach $\sim 59\%$, while MAT5 achieves $62.8\%$ with the best calibration ($20.6\%$ ECE). 
}

  \label{fig:gtsrb-rain-side}
  \vspace{-4mm}
\end{figure*}

\paragraph{Snow corruption.}
Figure~\ref{fig:gtsrb-snow-side} shows accuracy and calibration error under Snow corruption.  
Vanilla ERM and AT perform well on clean data (severity~0) but degrade rapidly, dropping below $70\%$ accuracy and above $30\%$ ECE at severity~5.  
AugMix is even less robust under snow, collapsing to near $55\%$ accuracy and $32\%$ ECE at severity~5.  
In contrast, model-based approaches maintain substantially higher performance across severities.  
MDA5 and MRT5 retain $76.1\%$ and $76.2\%$ accuracy at severity~5, with calibration error around $20\%$, clearly outperforming Vanilla and AT.  
MAT5 achieves comparable robustness ($75.8\%$) while consistently yielding the lowest ECE among single methods.  
Hybrid methods (MDAT5 and MRAT5) show modest gains at low to moderate severities: for example, MRAT5 reaches $97.2\%$ accuracy at severity~2 compared to $96.6\%$ for AT, and reduces ECE slightly ($4.1\%$ vs $4.7\%$).  
These improvements, while incremental, suggest that hybrids may help stabilize performance in the mid-severity regime, though at high severity (4–5) their advantage over MAT and MRT largely disappears.

\begin{tcolorbox}
\textbf{Takeaway 1:} Model-based methods dominate under Snow, with MAT5/MRT5 dominating across all severities and MDA5 achieving comparable performance with less computation complexity.
\end{tcolorbox}

\begin{figure*}[t]
  \centering
  \begin{subfigure}{\textwidth}
    \centering
    \includegraphics[width=\linewidth]{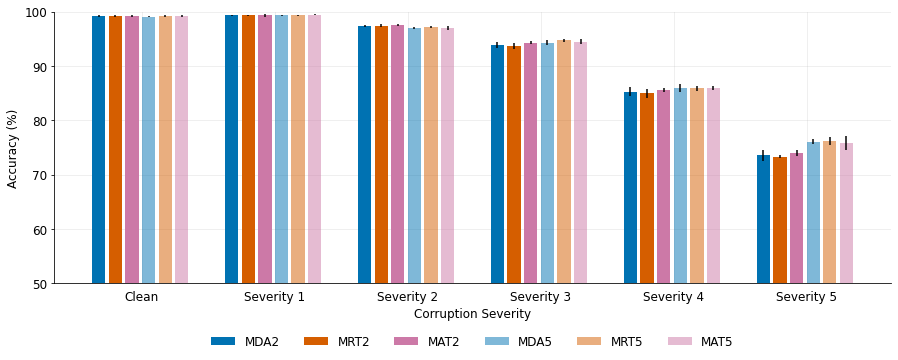}
    \caption{Accuracy (mean $\pm$ std) vs. severity.}
    \label{fig:snow-severity-comparisonacc}
  \end{subfigure}

  \vspace{1em} 

  \begin{subfigure}{\textwidth}
    \centering
    \includegraphics[width=\linewidth]{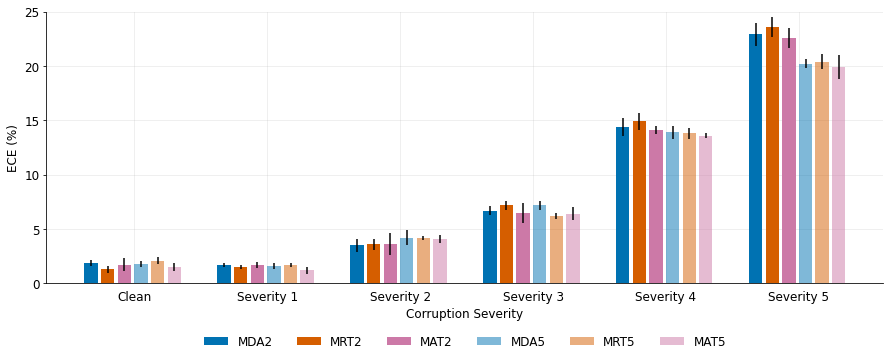}
    \caption{Accuracy (mean $\pm$ std) vs. severity.}
    \label{fig:snow-severitycomparisonece}
  \end{subfigure}

  \caption{\textbf{Comparing effect of Snow corruption training severity.} 
  We compare variants of MDA, MRT, and MAT trained with severity 2 and severity 5 augmentations. 
  Results show that severity 5 variants consistently achieve higher robustness at moderate to high corruption levels, while severity 2 variants perform slightly better under milder corruptions for Snow corruption.}
  \label{fig:snow-severitycomparison}
  \vspace{-7mm}
\end{figure*}
\paragraph{Rain corruption.}
Figure~\ref{fig:gtsrb-rain-side} presents results for Rain corruption, which is more challenging than Snow.
Vanilla and AT collapse sharply, reaching only $57.9\%$ and $54.0\%$ accuracy at severity~5, with calibration errors exceeding $30\%$.
AugMix again fails to generalize, falling to $29.7\%$ accuracy with $54.6\%$ ECE at severity~5.
Model-based methods substantially outperform these baselines: MDA5 ($59.2\%$), MRT5 ($59.5\%$), and MAT5 ($62.8\%$) each show clear gains in robustness, while their calibration remains $20$--$22\%$.
Among hybrids methods, MDAT5 and MRAT5 achieved comparable performance to MAT5 over severities~1--3 with more computational expense.
At severity~5, MAT5 achieves the highest robustness ($62.8\%$) and best calibration ($20.6\%$ ECE), slightly surpassing MRAT5 ($60.1\%$, $20.0\%$ ECE).
This highlights that for extreme corruptions, direct adversarial optimization in nuisance space (MAT) is more effective than hybrid initialization.

\begin{tcolorbox}
\textbf{Takeaway 2:} For Rain, MAT5 is the strongest across all severities surpassing hybrid methods in both accuracy and calibration but MDA5 achieved comparable performance with smaller computational complexity by a factor of $T$ (the number of PGD iterations).
\end{tcolorbox}

\vspace{-4mm}

\begin{figure*}[t]
  \centering
  \begin{subfigure}{\textwidth}
    \centering
    \includegraphics[width=\linewidth]{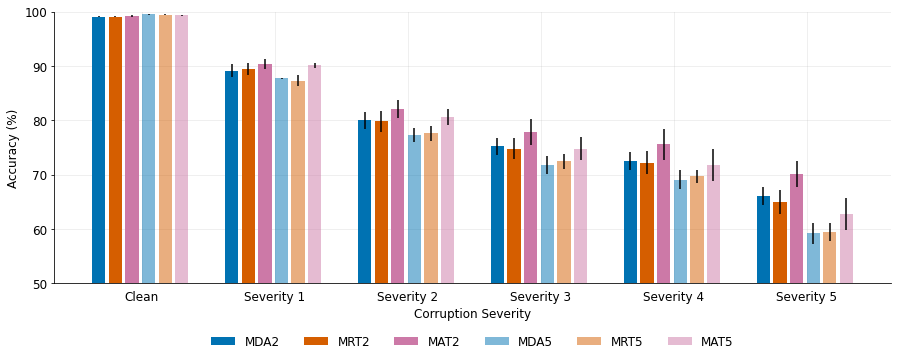}
    \caption{Accuracy (mean $\pm$ std) vs. severity.}
    \label{fig:rain-severity-comparisonacc}
  \end{subfigure}

  \vspace{1em} 

  \begin{subfigure}{\textwidth}
    \centering
    \includegraphics[width=\linewidth]{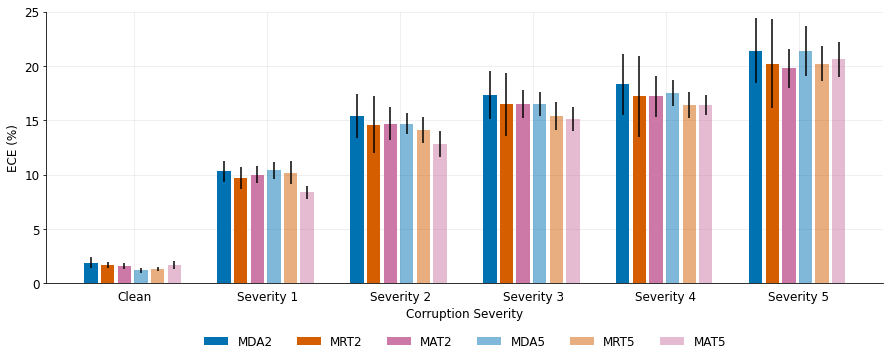}
    \caption{ECE (mean $\pm$ std) vs. severity.}
    \label{fig:rain-severity-comparisonece}
  \end{subfigure}

  \caption{\textbf{Comparing effect of Rain corruption training severity.} 
  We compare variants of MDA, MRT, and MAT trained with severity 2 and severity 5 augmentations. 
  Results show that severity 5 variants consistently achieve higher robustness at moderate to high corruption levels, while severity 2 variants perform slightly better under milder corruptions for Snow corruption.
  Interestingly MAT2 under Rain, outperformed all the methods trained with severity 5.}
  \label{fig:rain-severitycomparison}
  \vspace{-6mm}
\end{figure*}


\paragraph{Effect of training severity.}
We compare models trained with nuisance severity~2 (e.g., MDA2, MRT2, MAT2) against those trained with severity~5 (Figures~\ref{fig:snow-severitycomparison} and ~\ref{fig:rain-severitycomparison}).
At mild corruptions (severity~1--2), severity~2 training consistently yields lower calibration error, since the models are not overly specialized for extreme distortions.
For example, MAT2 achieves $3.6\%$ ECE at severity~2, compared to $4.1\%$ for MAT5 under Snow corruption (Figure~\ref{fig:snow-severitycomparisonece}).
At stronger corruptions (severity~4--5), severity~5 training provides higher robustness: MRT5 and MDAT5 both outperform their severity~2 counterparts by $2$--$3\%$ accuracy under Snow at severity~5.
Interestingly, under Rain, MAT2 generalizes best, reaching $70.2\%$ accuracy at severity~5, surpassing both MRT5 ($59.5\%$) and MDA5 ($59.2\%$) (Figure~\ref{fig:rain-severity-comparisonacc}).
This suggests that the optimal training severity depends on the corruption type: lower severities encourage better calibration, while higher severities improve robustness to strong distortions.

\begin{tcolorbox}
\textbf{Takeaway 3:} Training with severity~2 improves calibration and mild-corruption robustness, while severity~5 improves extreme-corruption robustness; MAT2 shows the best cross-severity generalization under Rain.
\end{tcolorbox}


\begin{table}[b]
\centering
\caption{\textbf{Welch's t-test ($\alpha = 0.05$) on the difference of means for Accuracy and Expected Calibration Error (ECE).} P-values and 95\% confidence intervals (CI) shown for the best MDA and MAT training methods on corruptions at severity 5. Results confirm that the difference in performance between the MDA and MAT algorithms is not statistically significant in three out of the four cases.}
\begin{tabular}{llllrrrr}
\toprule
\textbf{Corruption}   & \textbf{Method 1}     & \textbf{Method 2}     & \textbf{Metric} & \textbf{$\mu_1$} & $\mu_2$ & \textbf{95\% CI} & \textbf{$p$-value} \\
\midrule
\multirow{2}{*}{Snow} & \multirow{2}{*}{MDA5} & \multirow{2}{*}{MAT5} & Accuracy & 76.1 & 75.8 & $(-1.287,\ \ \  1.887)$ & 0.650 \\
                      &                       &                       & ECE & 20.2 & 19.9 & $(-1.042,\ \ \ 1.642)$    & 0.591              \\
\multirow{2}{*}{Rain} & \multirow{2}{*}{MDA2} & \multirow{2}{*}{MAT2} & Accuracy & 66.1 & 70.2 & $(-7.192, -1.007)$ & 0.016 \\
                      &                       &                       & ECE & 21.4 & 19.8 & $(-2.151,\ \ \ 5.351)$ & 0.343 \\
\bottomrule
\end{tabular}
\label{tab:stat_tests}
\end{table}

\vspace{-1mm}



\begin{wrapfigure}[3]{r}{0.40\textwidth} 
  \centering
  \includegraphics[width=\linewidth]{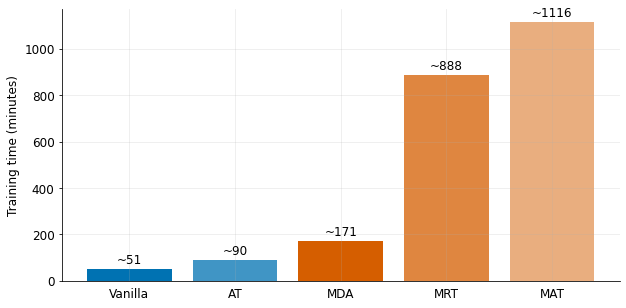}
  \caption{\textbf{Empirical training times (in minutes) for baseline methods.}
  Vanilla ERM is fastest, followed by AT and MDA. MRT and MAT are substantially
  more expensive due to repeated nuisance sampling and adversarial optimization.}
  \label{fig:train-times}
\end{wrapfigure}

\paragraph{Trade-offs and complexity.}
Table~\ref{tab:complexity} contextualizes these findings by comparing asymptotic training cost.
MDA and AugMix are $\Theta(1)$ but with larger constants than Vanilla due to augmentation overhead.
MRT scales linearly with the number of nuisance samples $k$, while MAT scales with the number of gradient iterations $T$ in nuisance space.
Hybrid variants inherit these costs: random-initialized hybrids are dominated by $\Theta(T)$ (like MAT), while worst-of-$k$ hybrids scale as $\Theta(k+T)$.
Empirical training times (Figure~\ref{fig:train-times}) confirm the expected ordering among the methods we measured: MAT $>$ MRT $>$ MDA $>$ PGD $>$ Vanilla.
Taken together with Figures~\ref{fig:gtsrb-snow-side} and~\ref{fig:gtsrb-rain-side}, this suggests a practical guideline:

\begin{itemize}
\item For \textbf{mild corruptions} (severities 1--3), MDA offers the best trade-off, substantially improving robustness and calibration over analytical baselines and achieving comparable results to MAT.
\item For \textbf{severe corruptions} (severities 4--5), MAT alone is most effective, outperforming hybrids at lower theoretical cost than MRAT.
\item When \textbf{efficiency is critical}, MDA offers a compromise: better robustness than Vanilla/AugMix but cheaper than MRT or MAT, without incurring a statistically significant drop in performance on Severity 5, as seen in Table \ref{tab:stat_tests}.
\end{itemize}

\paragraph{Summary.}
Across Snow and Rain corruptions, random analytical augmentations such as AugMix are insufficient. 
Model-based methods consistently improve both robustness and calibration, with MAT achieving the highest Mean Reciprocal Rank (MRR) across all methods, as seen in Table~\ref{tab:top5_mrr_combined}.
The hybrids methods achieved comparable performance to MAT, but at a higher computational expense.
Training severity also plays a role, severity~2 models yield better calibration and generalization to mild corruptions, whereas severity~5 models provide stronger robustness under extreme distortions.  
Finally, the complexity analysis reveals that MAT, while most robust across all severities, incurs the highest cost, while MDA remains the most efficient compromise, achieving comparable performance with $T$ less computational complexity.
\vspace{-2mm}

\vspace{4mm}
\begin{tcolorbox}
\textbf{Overall takeaway:} Robustness to natural corruptions is shaped both by the method and training severity. 
Model-based approaches (MAT, MRT, MDA) consistently outperform analytical augmentations. 
MAT achieves the strongest performance across all severities but with high computational cost and MDA offers the most efficient compromise. 
Training severity modulates outcomes: severity 2 favors calibration and generalization to mild corruptions, while severity 5 enhances robustness to extreme distortions.
\end{tcolorbox}

\section{Conclusion}

In this work, we conducted a comprehensive study of model-based and hybrid approaches for robustness to natural corruptions. 
We benchmarked a wide range of training methods on CURE-TSR under Snow and Rain corruptions, evaluating both accuracy and calibration across severities. 
Our analysis reveals clear trade-offs between robustness, calibration, and computational complexity. 
Overall, we draw three main conclusions:

\begin{tcolorbox}[colback=blue!5, colframe=blue!40, boxrule=0.5pt, arc=2pt, left=6pt, right=6pt, top=4pt, bottom=4pt]
1. \textbf{Model-based training improves robustness and calibration.} 
MAT, MRT, and MDA consistently outperform Vanilla, AT, and AugMix, retaining $75\%$ accuracy and $\sim 20\%$ ECE under severe corruptions.\\[0.5em]

2. \textbf{MDA provides the most efficient robustness–calibration trade-off.} 
While MAT achieves the highest performance across the different challenges, MDA strikes the best balance, offering strong robustness and calibration with substantially lower computational complexity without incurring a statistically significant drop in
performance on Severity 5.\\[0.5em]

3. \textbf{Training severity shapes robustness vs.~calibration.} 
Severity 2 training produces better calibration and mild-corruption generalization, while severity 5 training yields stronger robustness under extreme distortions; MAT generalizes best across severities for both corruptions, achieving the highest MRR.
\end{tcolorbox}

\begin{wraptable}[14]{r}{0.5\textwidth}
\vspace{-4mm}
\centering
\caption{\textbf{Mean Reciprocal Rank (MRR) for both Accuracy and Expected Calibration Error (ECE).} Training Methods shown correspond to the top 5 MRR scores (higher MRR value is better). MRR was calculated on results for Rain and Snow corruptions (severities 1–5) and the clean data.}
\begin{tabular}{lr}
\toprule
\textbf{Training Method} & \textbf{MRR} \\
\midrule
MAT5 & 0.506 \\
MRAT2 & 0.475 \\
MAT2 & 0.435 \\
MDAT2 & 0.356 \\
MRT5 & 0.339 \\
\bottomrule
\end{tabular}
\label{tab:top5_mrr_combined}
\end{wraptable}
\textbf{Limitations.} 
Our study is limited to a single dataset (CURE-TSR) and two corruption types (Snow, Rain). 
We also rely on a simple CNN backbone and a MUNIT-style variation model, which may not reflect behavior under larger architectures or other generative models. 
Future work should extend this evaluation to more datasets, architectures, and corruption types to test the generality of these findings.

\textbf{Moving forward.} 
Our results indicate that combining nuisance coverage and adversarial refinement is a promising direction for robustness to natural corruptions. 
Three avenues stand out: 
(1) extending hybrids to additional modalities (e.g., audio, video, multimodal perception); 
(2) developing more efficient approximations of MRT and MRAT to reduce computational cost; 
and (3) scaling to larger backbones and datasets to examine whether these trade-offs persist in high-capacity regimes.

\begin{ack}
DISTRIBUTION STATEMENT A. Approved for public release. Distribution is unlimited.

This material is based upon work supported by the Under Secretary of Defense for Research and Engineering under Air Force Contract No. FA8702-15-D-0001. Any opinions, findings, conclusions or recommendations expressed in this material are those of the author(s) and do not necessarily reflect the views of the Under Secretary of Defense for Research and Engineering.

\copyright \ 2025 Massachusetts Institute of Technology.

Delivered to the U.S. Government with Unlimited Rights, as defined in DFARS Part 252.227-7013 or 7014 (Feb 2014). Notwithstanding any copyright notice, U.S. Government rights in this work are defined by DFARS 252.227-7013 or DFARS 252.227-7014 as detailed above. Use of this work other than as specifically authorized by the U.S. Government may violate any copyrights that exist in this work.

\end{ack}

\bibliographystyle{plain}
\bibliography{references}

\end{document}